\documentclass[10pt,journal]{IEEEtran}

\ifCLASSOPTIONcompsoc
  \usepackage[nocompress]{cite}
\else
  \usepackage{cite}
\fi

\usepackage{times}
\usepackage{graphicx}
\usepackage{amsmath}
\usepackage{amsthm}
\usepackage{amssymb}
\usepackage{booktabs}
\usepackage{multirow}

\usepackage{algorithm}
\usepackage{algpseudocode}

\usepackage{subcaption}
\usepackage{adjustbox}
\usepackage{svg}

\usepackage{pifont}%

\usepackage[pagebackref=true,breaklinks=true,colorlinks,bookmarks=false]{hyperref}
\usepackage[capitalise, noabbrev]{cleveref}
\usepackage{orcidlink}

\newcommand*{\bfx}{\mathbf{x}}
\newcommand*{\bfz}{\mathbf{z}}

\newcommand{\etal}{\textit{et al}.}
\newcommand{\ie}{\textit{i}.\textit{e}.}
\newcommand{\eg}{\textit{e}.\textit{g}.}

\begin{document}

\title{Fast-DiM: Towards Fast Diffusion Morphs}

\author{Zander W.~Blasingame~%
and Chen~Liu\\
Department of Electrical and Computer Engineering\\
Clarkson University, Potsdam, New York, USA\\
\texttt{\{blasinzw; cliu\}@clarkson.edu}
}

\markboth{\tiny{THIS WORK HAS BEEN SUBMITTED TO THE IEEE FOR POSSIBLE PUBLICATION. COPYRIGHT MAY BE TRANSFERRED WITHOUT NOTICE, AFTER WHICH THIS VERSION MAY NO LONGER BE ACCESSIBLE.}}%
{}

\IEEEtitleabstractindextext{%
\begin{abstract}
Diffusion Morphs (DiM) are a recent state-of-the-art method for creating high quality face morphs; however, they require a high number of network function evaluations (NFE) to create the morphs.
We propose a new DiM pipeline, Fast-DiM, which can create morphs of a similar quality but with fewer NFE.
We investigate the ODE solvers used to solve the Probability Flow ODE and the impact they have on the the creation of face morphs.
Additionally, we employ an alternative method for encoding images into the latent space of the Diffusion model by solving the Probability Flow ODE as time runs forwards.
Our experiments show that we can reduce the NFE by upwards of 85\% in the encoding process while experiencing only 1.6\% reduction in Mated Morph Presentation Match Rate (MMPMR).
Likewise, we showed we could cut NFE, in the sampling process, in half with only a maximal reduction of 0.23\% in MMPMR.
\end{abstract}

}

\maketitle

\IEEEdisplaynontitleabstractindextext

\IEEEpeerreviewmaketitle

\section{Introduction}
\begin{figure}[t]
    \centering
    \begin{subfigure}{0.16\textwidth}
        \includegraphics[width=0.98\textwidth]{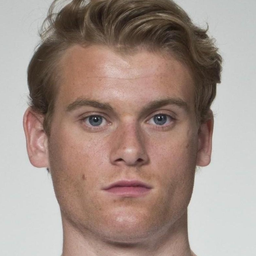}
        \caption{Identity $a$}
    \end{subfigure}%
    \begin{subfigure}{0.16\textwidth}
        \includegraphics[width=0.98\textwidth]{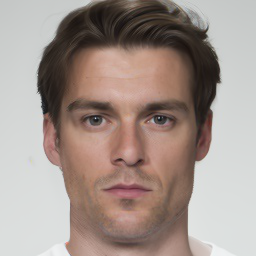}
        \caption{Morphed image}
    \end{subfigure}%
    \begin{subfigure}{0.16\textwidth}
        \includegraphics[width=0.98\textwidth]{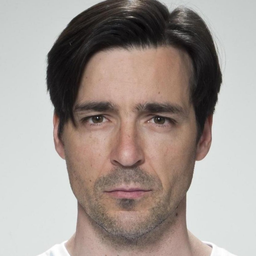}
        \caption{Identity $b$}
    \end{subfigure}
    \caption{Example of a morph created using DiM~\cite{blasingame2023leveraging}. Samples are from the FRLL dataset~\cite{frll}.}
    \label{fig:morph_ex}
\end{figure}

\IEEEPARstart{F}{ace} recognition (FR) systems are a common biometric modality used for identity verification across a diverse range of applications, from simple tasks such as unlocking a smart phone to official businesses such as banking, e-commerce, and law enforcement.  
Unfortunately, while FR systems can reach excellent performance with low false rejection and acceptance rates, they are uniquely vulnerable to a new class of attacks, that is, the face morphing attack
~\cite{Blasingame2021LeveragingAL}.
Face morphing attacks aim to compromise one of the most fundamental properties of biometric security, \ie, the one-to-one mapping from biometric data to the associated identity.
To achieve this the attacker creates a morphed face which contains biometric data of both identities.
Then \textit{one} morphed image, when presented, forces the FR system to register a match with \textit{two} disjoint identities, violating this fundamental principle, see~\cref{fig:morph_ex} for an example.

Face morphing attacks, thus, pose a significant threat towards FR systems.
One notable affected area by this attack is the e-passport, wherein the applicant submits a passport photo either in digital or printed format.
This is particularly relevant for countries where e-passports are used for both issuance and renewal of documents.
Critically, an adversary who is blacklisted from accessing a certain system, such as e-passport, can create a morph to gain access as a non-blacklisted individual.

In response to the severity of face morphing attacks, an abundance of algorithms have been developed to identify these attacks~\cite{Blasingame2021LeveragingAL}.
There are two broad classes of Morphing Attack Detection (MAD) algorithms based on the scenario in which they operate.
The first scenario is where the MAD algorithm is only shown a single image and tasked with deciding if the particular image is a morphed image or a bona fide image~\cite{mipgan}.
Algorithms which solve this problem are known as Single image-based MAD (S-MAD) algorithms.
The second scenario is where the MAD algorithm is presented two images, of which one image is verified to be a bona fide image, \eg, through live capture, and the other is the unknown image that the model is tasked to classify.
Algorithms which solve this problem are known as Differential MAD (D-MAD) algorithms~\cite{mipgan}.
By construction the S-MAD problem is much more difficult that the D-MAD problem, as the D-MAD algorithm has the guaranteed bona fide image to compare against, whereas the S-MAD problem offers no such luxury~\cite{Blasingame2021LeveragingAL}.

A plethora of morphing attacks have been developed, for the purposes of this work we broadly categorize them into two categories: landmark-based morphing attacks and representation-based morphing attacks~\cite{blasingame2023leveraging}.
Landmark-based morphing attacks use local features to create the morphed image by warping and aligning the landmarks within each face then followed by pixel-wise compositing.
Landmark-based attacks have been shown to be highly effective against FR systems~\cite{syn-mad22}.
In contrast, representation-based morphing attacks use a machine learning model to embed the original bona fide faces into a representation space which are then combined to produce a new representation that contains information from both identities.
This new representation is then used by a generative model to construct the morphed image.
Recently, there has been an explosion of work exploring deep-learning based face morphing using generative models like Generative Adversarial Networks (GANs)~\cite{mipgan}.
Currently, FR systems seem especially vulnerable to landmark-based attacks~\cite{mipgan, blasingame2023leveraging}; however, landmark-based attacks are prone to more noticeable artefacts than representation-based attacks~\cite{blasingame2023leveraging}.

Recent work has shown that Diffusion Morphs (DiM) can achieve state-of-the-art performance rivaling that of GAN-based methods~\cite{blasingame2023leveraging}.
However, DiM requires a high number of Network Function Evaluations (NFE), incurring great computation demand and complexity. This renders the integration of further techniques, like identity-based optimization, much more difficult.
To draw samples from Diffusion models, an initial image of white noise is deployed and the Probability Flow Ordinary Differential Equation (PF-ODE) is solved as time runs backwards~\cite{song2021scorebased}.
However, for DiM models an additional encoding step from the original image back to noise is needed, which also uses many NFE to calculate this encoding~\cite{diffae,blasingame2023leveraging}.
We posit that this encoding step is identical to solving the PF-ODE as time runs forwards and propose to use an additional ODE solver to accomplish this encoding.
Additionally, we propose to use ODE solvers with faster convergence guarantees in solving the PF-ODE, in lieu of the slower ODE solver used by previous work~\cite{blasingame2023leveraging}.
We then study the impact of these design choice on the application of face morphing.
We summarize our contributions in this work as follows:
\begin{enumerate}
    \item We propose a novel morphing method named Fast-DiM which can achieve similar performance to DiM but with greatly reduced number of NFE.
    \item We perform an extensive study on the impact of the ODE solvers on DiMs. %
    \item We compare our method to state-of-the-art morphing attacks via a vulnerability and detectability study.
    \item To the best of our knowledge, we are the first to study the impact of ODE solvers for the Probability Flow ODE as time runs forward on autoencoding tasks.
\end{enumerate}

\section{Prior Work}
A na\"ive but simple approach to construct face morphs is to simply take a pixel-wise average of the two images. Unsurprisingly, this approach often yields significant artefacts.
These artefacts are especially apparent when the images are not aligned, resulting in strange deformations, \eg, the mouth of one subject overlaps with the nose of another or the morphed image containing four eyes.
A simple remedy to this problem is to align the faces so that the key landmarks of the faces overlap, \ie, the nose of subject one aligns with the nose of subject two and so forth with each landmark.
The approaches which use this system of warping and aligning the images so the landmarks overlap for each face before taking a pixel-wise average to construct the morph are known as Landmark-based morphs.
These Landmark-based morphs often exhibit artefacts outside the core area of the face, enabling easy detection of the morphing attack through simple visual inspection or with MAD algorithms~\cite{blasingame2023leveraging, syn-mad22}.

Later work explored the use of deep generative models to create face morphs.
The key idea in this approach is to perform the morphing at the representation-level rather than at the pixel-level.
Initial work in this direction pursued the use of Generative Adversarial Networks (GANs) for this purpose.
GANs train a generator network through an adversarial strategy.
This generator network maps latent vectors from, a typically low dimensional manifold, to the image space.
Now to enable face morphing attacks, an encoding strategy which can embed images into the latent space of the generator is needed.
This encoding strategy could be an additional encoding network, or something else like optimization with Stochastic Gradient Descent (SGD).
Additionally, this encoding strategy needs to have low distortion on the inversion, \ie, an image that is encoded and then mapped back to the image space via the generator should be ``very close'' to the original image.
Using this encoding strategy, the latent representations for two identities are then averaged to produce a new latent representation, \ie, the morphed latent.
This morphed latent is then used as the input to the generator which maps the morphed latent back into the image space, yielding the morphed image.

The MIPGAN model by Zhang~\etal~\cite{mipgan} proposes an extension on GAN-based approach by adding an identity-based loss function derived from an FR system and using it to optimize the morph creation process.
Two bona fide images are embedded into the latent space using an encoding network which predicts the latents from the original images.
The two latent representations from this procedure are denoted as $z_a, z_b$ for subjects $a$ and $b$, separately.
The morphed latent representation is initially constructed as a linear interpolation between these two latents, \ie, $z_{ab} = \frac 12(z_a + z_b)$.
This initial representation is used as the starting point for a second optimization procedure, wherein the optimal morphed latent is found such that the  output of generator network $G(z_{ab})$ is minimized with respect to a loss function which measures the similarity between the morphed face and two bona fide faces via an FR system and additional perceptual loss metrics.
At the end of this optimization procedure, a morphed face should have been found which fools the FR system used to guide the optimization procedure.

Blasingame~\etal~\cite{blasingame2023leveraging} propose DiM, a novel face morphing approach which uses Diffusion Autoencoders~\cite{diffae} to construct morphed faces.
Unlike GANs, which learn a mapping from the latent space to the image space, Diffusion models consider a Stochastic Differential Equation (SDE) which perturbs the initial image distribution into an isotropic Gaussian on the image space over time, given by the It\^o SDE
\begin{equation}
    \label{eq:sde-diffusion}
    \text d \bfx_t =f(t)\bfx_t\text dt + g(t)\text d\mathbf{w}_t
\end{equation}
where $t \in [0, T]$, $f, g$ are the drift and diffusion coefficients and $\mathbf{w}_t$ is the Brownian motion.
Let $\alpha_t, \sigma_t$ be the noise schedule of the diffusion process where  $\alpha_t$ denotes how much of the original image $\bfx_0$ is present at time $t$ and $\sigma_t$ denotes how much noise is present, such that at any time $t$
\begin{equation}
    \label{eq:forward_eq}
    \bfx_t = \alpha_t \bfx_0 + \sigma_t \epsilon
\end{equation}
for some Gaussian noise $\epsilon$.
Then the drift and diffusion coefficients are
\begin{equation}
    f(t) = \frac{\text d \log \alpha_t}{\text dt}, g^2(t) = \frac{\text d\sigma_t^2}{\text d t} - 2 \frac{\text d \log \alpha_t}{\text dt}\sigma_t^2
\end{equation}
Song~\etal~\cite{song2021scorebased} show that there exists a reverse Ordinary Differential Equation (ODE) known as the \textit{Probability Flow ODE} (PF-ODE) with the same marginals as $p_t(\bfx_t)$ exists with the form
\begin{equation}
    \label{eq:pf_ode}
    \frac{\text d\bfx_t}{\text dt} =  f(t)\bfx_t- g^2(t) \nabla_\bfx \log p_t(\bfx_t)
\end{equation}
as time flows backwards from $T$ to $0$ where $\nabla_\bfx \log p_t(\bfx_t)$ is called the score function.
Diffusion models learn to model this score function with a neural net, often a U-Net, $\boldsymbol\epsilon_\theta(\bfx_t) \approx -\sigma_t\nabla_\bfx \log p_t(\bfx_t)$.
By using the learned score function, a wide array of numerical ODE solvers can be deployed to solve the PF-ODE, enabling sampling of the data distribution $p_0(\bfx_0)$ by drawing an initial condition $\bfx_T \sim p_T(\bfx_T)$ from the isotropic Gaussian and running the ODE solver.

The Diffusion Autoencoder model consists of a conditioned noise prediction U-Net $\boldsymbol{\epsilon}_\theta: (\bfx_t, \bfz, t) \mapsto \hat\epsilon$ and encoder network $\mathcal{E}:\bfx_0 \mapsto \bfz$ which learns the latent representation for an image~\cite{diffae}.
This model uses the deterministic version of the Denoising Diffusion Implicit Model (DDIM) solver $\phi_{t_i}: (\bfx_{t_i}, \hat\epsilon) \mapsto \bfx_{t_{i-1}}$ where $\{t_i\}_{i=1}^N \subseteq [0, T]$ is the time schedule used for sampling with $N$ inference steps.
Additionally, the deterministic DDIM solver is reversed to introduce the ``stochastic encoder'' $\phi_{t_i}^{-}: (\bfx_{t_i}, \hat\epsilon) \mapsto \bfx_{t_{i+1}}$.
In this work we refer to this ``stochastic encoder'' as a DiffAE forward solver, since it has a similar objective to solving~\cref{eq:pf_ode} as time runs forwards from $0$ to $T$.

\begin{algorithm}[h]
    \caption{The DiM Algorithm.}
    \label{alg:ddim_morphing}
    \begin{algorithmic}
        \State $\bfz_a \gets \mathcal{E}(\bfx_0^{(a)})$
        \State $\bfz_b \gets \mathcal{E}(\bfx_0^{(b)})$
        \For {$i \gets 1, 2, \ldots N$}
            \State $\bfx_{t_{i+1}}^{(a)} \gets \phi_{t_i}^{-}(\bfx_{t_i}^{(a)}, \boldsymbol{\epsilon}_\theta(\bfx_{t_i}^{(a)}, \bfz_a, t_i))$ 
            \State $\bfx_{t_{i+1}}^{(b)} \gets \phi_{t_i}^{-}(\bfx_{t_i}^{(b)}, \boldsymbol{\epsilon}_\theta(\bfx_{t_i}^{(b)}, \bfz_b, t_i))$ 
        \EndFor
        \State $\bfx_T^{(ab)} \gets \textrm{slerp}(\bfx_T^{(a)}, \bfx_T^{(b)}, 0.5)$
        \State $\bfz_{ab} \gets \frac 12(\bfz_a + \bfz_b)$
        \For {$i \gets N, N - 1, \ldots 2$}
            \State $\bfx_{t_{i-1}}^{(ab)} \gets \phi_{t_i}(\bfx_{t_i}^{(ab)}, \boldsymbol{\epsilon}_\theta(\bfx_{t_i}^{(ab)}, \bfz_{ab}, t_i))$ 
        \EndFor
        \State \textbf{return} $\bfx_0^{(ab)}$
    \end{algorithmic}
\end{algorithm}

Let $\bfx_0^{(a)}, \bfx_0^{(b)}$ denote face images of two \textit{bona fide} subjects, $a$ and $b$, separately.
The DiM morphing procedure is outlined in~\cref{alg:ddim_morphing}.
At a high level this approach encodes the bona fide images into stochastic latent codes $\bfx_T^{(a)}$ and $\bfx_T^{(b)}$ by running the reverse DDIM solver.
These stochastic latent codes are then morphed using spherical interpolation\footnote{For a vector space $V$ and two vectors $u, v \in V$, the spherical interpolation by a factor of $\gamma$ is given as
\begin{equation*}
    \textrm{slerp}(u, v; \gamma) = \frac{\sin((1 - \gamma) \theta)}{\sin \theta}u + \frac{\sin(\gamma \theta)}{\sin \theta}v
\end{equation*}
where $\theta = \frac{\arccos(u \cdot v)}{\|u\| \, \|v\|}$.} to give the morphed stochastic latent code $\bfx_T^{(ab)}$.
The semantic latent codes are averaged to obtain $\bfz_{ab}$.
These morphed latent codes are then used with the DDIM solver to generate the morphed image $\bfx_0^{(ab)}$.
This approach is labeled variant A in~\cite{blasingame2023leveraging}, while the authors also recommend another approach called variant C which uses an additional ``pre-morph'' stage to alter the bona fide images before encoding.
We call these two approaches DiM-A and DiM-C, respectively.
In this work we primarily investigate and improve upon two aspects of the DiM framework, which are the mechanism for encoding $\bfx_0$ into $\bfx_T$, \ie, the forward ODE solver $\phi_t^{-}$, and the mechanism for generating the morphed image from $\bfx_T^{(ab)}$, \ie, the PF-ODE solver $\phi_t$.

\section{Experimental Setup}
Here we outline the structure of our experiments, giving details on how we compare our proposed morphing attack Fast-DiM against other methods.
We explain the dataset we use for evaluation, the FR systems we evaluate against, and the metrics we use to assess the vulnerability of the FR systems to morphing attacks.

\subsection{Dataset}

To evaluate the effectiveness of the morphing algorithms explored in this paper, we use the SYN-MAD 2022\footnote{\url{https://github.com/marcohuber/SYN-MAD-2022}} competition dataset~\cite{syn-mad22}.
The SYN-MAD 2022 dataset consists of pairs of identities used for face morphing from the Face Research Lab London (FRLL) dataset~\cite{frll}.
The FRLL dataset consists of high-quality samples of 102 different individuals with two images per subject, one of a ``neutral'' expression and the other of a ``smiling'' expression.
The ElasticFace~\cite{elasticface} FR system was used to calculate the embedding of all the frontal images of the FRLL dataset.
Once the embedding were calculated, the top 250 most similar pairs for each gender, in terms of cosine similarity, were selected~\cite{syn-mad22}.
These 500 pairs are used to create the morphed images.
In this work we use only the 500 pairs of ``neutral'' images when creating and evaluating the morphs.

The SYN-MAD 2022 creates morphs with three landmark-based approaches and two GAN-based approaches.
These are the open-source OpenCV\footnote{\url{https://learnopencv.com/face-morph-using-opencv-cpp-python/}}, commercial-of-the-shelf (COTS) FaceMorpher\footnote{\url{https://www.luxand.com/facemorpher/}}, and online-tool Webmorph\footnote{\url{https://webmorph.org/}} landmark-based morphing algorithms.
Note the FaceMorpher from SYN-MAD 2022 is not the same as the FaceMorpher from~\cite{blasingame2023leveraging} which is another landmark-based open-source face morphing algorithm of the same name.
The two GAN-based algorithms are the MIPGAN-I and MIPGAN-II models from~\cite{mipgan}.
We run the DiM algorithm on this dataset using variants A and C from~\cite{blasingame2023leveraging} on the same 500 pairs to evaluate against previous Diffusion-based work.
The OpenCV morphs from the SYN-MAD 2022 dataset consist of only 489 morphs due to technical issues with the other 11 morphs.
To ensure a fair comparison all evaluation is done on this subset of the SYN-MAD 2022 dataset.

We align all the bona fide images from FRLL and the landmark-based morphs are aligned and cropped to the face using the dlib library based on the alignment pre-processing used to create FFHQ dataset~\cite{stylegan}.
As the MIPGAN and DiM algorithms use the alignment script when creating their morphs, it was not necessary to re-run the alignment script on the morphs created from these algorithms.

\begin{table*}[t]
    \centering
    \caption{Comparing the design choices for DiM in Blasingame~\etal~\cite{blasingame2023leveraging} versus our modifications.}
    \resizebox{\textwidth}{!}{%
    \begin{tabular}{lll}
    \toprule
    & \textbf{Design Choice in Blasingame~\etal~\cite{blasingame2023leveraging}} & \textbf{Our Modifications}\\
    \midrule
    ODE Solver & DDIM & DPM++ 2M\\
    \hline
    Forward ODE Solver &
    $\begin{aligned}
    \bfx_{t_{i+1}} &= \frac{\sigma_{t_{i+1}}}{\sigma_{t_{i}}}\bigg(\bfx_{t_{i}} + \alpha_{t_i} (e^{h_{i+1}}-1)\bfx_\theta(\bfx_{t_{i}}, \bfz, t_{i})\bigg)
    \end{aligned}$
    &
    $\begin{aligned}
         \bfx_{t_{i+1}} &= \frac{\sigma_{t_{i+1}}}{\sigma_{t_i}}\bfx_{t_i} - \alpha_{t_{i+1}}(e^{-h_{i+1}} - 1)\bfx_\theta(\bfx_{t_i}, \bfz, t_{i})
    \end{aligned}$\\
    \hline
    Parameters & $N = 100, N_F = 250$ & $N = 50, N_F = 100$\\
    \hline
    Shared Parameters & \multicolumn{2}{c}{$\beta_i = \frac{(i-1) (\beta_{max} - \beta_{min})}{N_T} + \beta_{min}$ where $i \in [\![1, N_T]\!], N_T = 1000, \beta_{min} = 0.0001, \beta_{max} = 0.02$}\\
    & \multicolumn{2}{c}{$\alpha_i^2 = \prod_{n=1}^i (1 - \beta_n), \sigma_i = \sqrt{1 - \alpha_i^2}, \lambda_i = \log \frac{\alpha_i}{\sigma_i}, h_i = \lambda_{t_i} - \lambda_{t_{i-1}}$}\\
    \bottomrule
    \end{tabular}
    }
    \label{tab:design_choices}
\end{table*}

\subsection{Face Recognition Systems}

Three publicly available FR systems were used to evaluate the effectiveness of the face morphing attacks. In particular, the ArcFace\footnote{\url{https://github.com/deepinsight/insightface}}~\cite{deng2019arcface}, AdaFace\footnote{\url{https://github.com/mk-minchul/AdaFace}}~\cite{adaface}, and ElasticFace\footnote{\url{https://github.com/fdbtrs/ElasticFace}}~\cite{elasticface} FR systems were used.
All systems convert the input image into an embedding within a high-dimensional vector space. The distance between the embeddings of the probe and target images are then compared to determine if the probe image belongs to the same identity as the target image.
If this distance is sufficiently ``small'', the probe image is said to belong to the same identity as the target image.

The ArcFace system is based on the Improved ResNet (IResNet-100) architecture trained on the MS1M-RetinaFace dataset\footnote{\label{fnote:datasets}\url{https://github.com/deepinsight/insightface}}.
The IResNet architecture is able to improve from the baseline ResNet architecture without increasing the number of parameters or computational costs.
The ArcFace system uses an additive angular margin loss which aims to enforce intra-class compactness and inter-class distance.

ElasticFace~\cite{elasticface} builds upon the work of ArcFace by relaxing the fixed penalty margin used by ArcFace and proposes to use an elastic penalty margin loss.
These improvements allow ElasticFace to achieve state-of-the-art performance.
The ElasticFace system used in this paper is based on the IResNet-100 architecture trained on the MS1M-ArcFace dataset\footnote{See~\cref{fnote:datasets}.}.

AdaFace uses an adaptive margin loss by approximating the image quality with feature norms~\cite{adaface}. This approximation of image quality is used to give less weight to misclassified samples during training that have ``low'' quality. This improvement to the loss allows the system to achieve state-of-the-art recognition performance.
The AdaFace system used in this paper is based on the IResNet-100 architecture trained on the MS1M-ArcFace dataset.

All three FR systems require an input of $112 \times 112$ pixels. Every image is resized such that the shortest side of the image is 112 pixels long.
The resulting image is then cropped into a $112 \times 112$ pixel grid.
The image is then normalized so the pixels take values in $[-1, 1]$.
Lastly, the AdaFace system was trained on BGR images so the image tensor is shuffled from the RGB to the BGR format for the AdaFace system.

\begin{figure*}[t]
    \centering
    \includegraphics[width=0.8\textwidth]{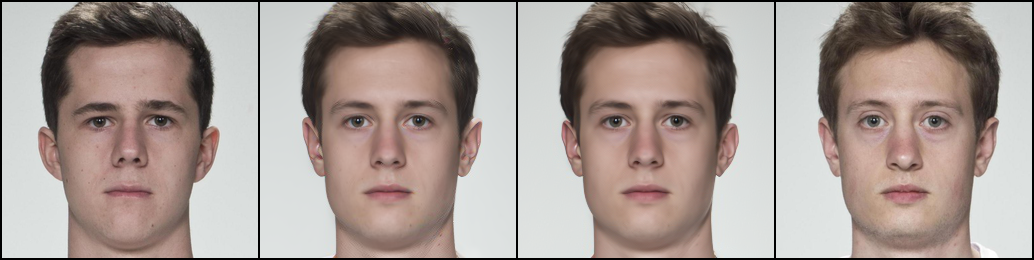}
    \caption{From left to right: identity $a$, morph generated with DDIM $(N = 100)$, morph generated with DPM++ 2M $(N = 20)$, identity $b$.}
    \label{fig:ddim-vs-dpm}
\end{figure*}

\subsection{Metrics}

The Mated Morph Presentation Match Rate (MMPMR) metric is widely used as a measure of vulnerability of FR systems when facing morphing attacks.
The MMPMR metric proposed by Scherhag~\etal~\cite{mmpmr} is defined as
\begin{equation}
M(\delta) = \frac{1}{M}\sum_{m=1}^{M} \left\{ \left[ \min_{n \in \{1,\ldots,N_m\}} S_m^n \right] > \delta \right\}
\end{equation}
where $S_m^n$ is the similarity score of the $n$-th subject of morph $m$, $M$ is the total number of morphed images, $\delta$ is the verification threshold, and $N_m$ is the total number of subjects contributing to morph $m$.
In practice the verification threshold $\delta$ is set to achieve a pre-specified False Match Rate (FMR) for the given FR system.
The similarity score, or conversely distance score, is a measure of the difference between the embeddings for the morphed image and bona fide image.
For our experiments we use the cosine distance to measure the distance between embeddings.

Morphing Attack Potential (MAP) is an extension on the MMPMR metric proposed by Ferrara~\etal~\cite{map_metric} which aims to provide a more comprehensive assessment of the risk a particular morphing attack poses to FR systems.
The MAP metric is a matrix such that $\text{MAP}[r, c]$ denotes the proportion of morphed images that successfully trigger a match decision against at least $r$ attempts for each contributing subject by at least $c$ of the FR systems~\cite{map_metric}.
Since the SYN-MAD 2022 only has one probe image per subject, as we exclude the bona fide image used in the creation of the morph, we simply report $\text{MAP}[1, c]$ which still provides insight into the generality of a morphing attack.

Additionally, we use the Learned Perceptual Image Patch Similarity (LPIPS)~\cite{LPIPS} metric to assess the similarity between images.
LPIPS computes the similarity between the activations of two images for some neural network.
This measure has been shown to correlate well with human assessment of image similarity~\cite{LPIPS}.
We use the VGG network as the backbone for the LPIPS metric in our experiments.

\section{Fast-DiM}
We present our novel morphing algorithm, Fast-DiM, as a series of design considerations and changes from the original DiM model, which we summarize in~\cref{tab:design_choices}.
In our initial design exploration we found that DiM-A outperformed DiM-C slightly, in contrast with the recommendation of~\cite{blasingame2023leveraging}.
This could be in part due to differences in FR systems as we chose a more modern set of FR systems to evaluate on.
Nevertheless, because of this initial strength of DiM-A over DiM-C in our own testing, we start from developing our Fast-DiM model from DiM-A.
For our experiments we measure the MMPMR values on the SYN-MAD 2022 dataset across the ArcFace, ElasticFace, and AdaFace FR systems in addition to reporting the NFE for each model.
Note, the False Match Rate (FMR) is set at 0.1\% for each FR system when reporting the MMPMR.

\subsection{The ODE Solver}
\label{sec:ode_solver}
We begin by swapping out the DDIM ODE solver for a faster ODE solver as
Preechakul~\etal~\cite{diffae} recommend 100 iterations using the DDIM solver.
Thankfully, much research has been done on developing numerical ODE solvers which can solve the PF-ODE, see~\cref{eq:pf_ode}, with fewer steps.
Lu~\etal~\cite{lu2023dpmsolver} proposed the DPM++ solver which is a high-order multi-step ODE solver developed specifically for solving the PF-ODE.
We implement the DPM++ 2M solver to work with the U-Net from the Diffusion Autoencoder model to allow for faster sampling.
DPM++ 2M is a second-order multi-step solver which achieves state-of-the-art performance compared to other ODE solvers~\cite{lu2023dpmsolver}.
Following the algorithm outlined in~\cref{alg:ddim_morphing}, the original DDIM solver $\phi_t$ is swapped out with the DPM++ 2M solver.
\cref{fig:ddim-vs-dpm}\footnote{Note that for illustrative purposes our figures use the DiM-C variant as it has less visible artefacts.} provides an illustration of two morphs, one generated with the original DDIM solver with $N = 100$ steps and the other generated using the DPM++ 2M solver with $N = 20$ steps.
Remarkably, there is little difference upon visual inspection, providing great hope that the DPM++ 2M solver can simply be used in lieu of the DDIM solver.

\begin{table}[h]

    \caption{Impact of ODE Solver on the DiM-A algorithm.}
    \centering
    \begin{tabular}{lrrrr}
    \toprule
        && \multicolumn{3}{c}{\textbf{MMPMR}($\uparrow$)}\\
        \cmidrule(lr){3-5}
     \textbf{ODE Solver}                                 &   \textbf{NFE}($\downarrow$) & \textbf{AdaFace} &   \textbf{ArcFace} &  \textbf{ElasticFace} \\
    \midrule
      DDIM                                           &100&  92.23 &               90.18 &                   93.05 \\
     DPM++ 2M  &50&               92.02 &               90.18 &                   93.05 \\
     DPM++ 2M   &20&               91.62 &               89.98 &                   93.25 \\
    \bottomrule
    \end{tabular}
    \label{tab:ode_solver}
\end{table}

The impact of NFE and the choice of ODE solver on the potency of a morph is outlined in~\cref{tab:ode_solver}.
Noticeably, there is little to no reduction in MMPMR across all three FR systems when using DPM++ 2M with $N = 50$ steps versus the DDIM solver with $N = 100$ steps, meaning that switching to this solver is a straightforward improvement in NFE while essentially sacrificing no morphing performance.
However, there is a slight reduction in MMPMR values when using $N=20$ steps.
As such we recommend the DPM++ 2M solver with $N = 50$ as it reduces NFE by half while requiring no compromise in morphing performance.

\begin{figure}
    \centering
    \includegraphics[width=0.485\textwidth]{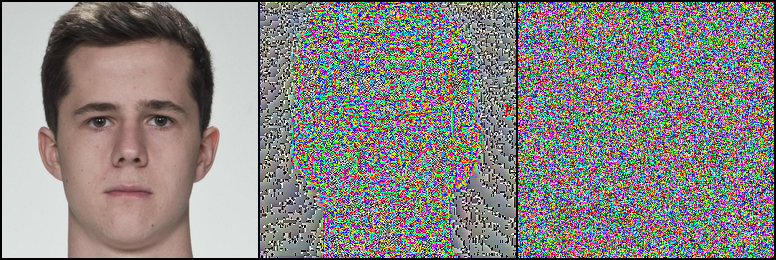}
    \includegraphics[width=0.485\textwidth]{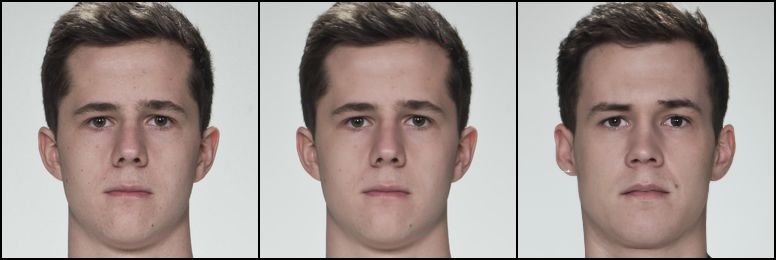}
    \caption{From top left to lower right: original image, output from the DiffAE forward solver, white noise, original image, DDIM sampled image from DiffAE approach, DDIM sampled image from pure white noise.}
    
    \label{fig:ddim-encode}
\end{figure}

\begin{figure*}[t]
    \centering
    \includegraphics[width=\textwidth]{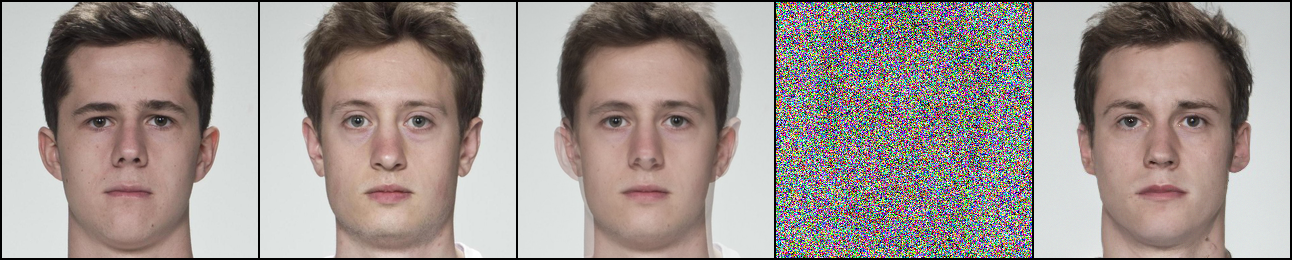}
    \caption{From left to right: identity $a$, identity $b$, pixel-wise averaged image, noisy image, final morphed image.}
    \label{fig:dpm-partial-noise}
\end{figure*}

\subsection{Noise Injection}
\label{sec:noise_inject}
Next we ask if it is even necessary to use the forward ODE solver, $\phi_t^-$, or if we can simply start the morph from some random noise, $\bfx_T \sim \mathcal{N}(\mathbf{0}, \mathbf{I})$.
As $\bfz$ is intended to contain all the semantic details, we wonder if all the information necessary to create an image which would fool an FR system was contained within $\bfz$ or if it was also contained in $\bfx_T$.
In~\cref{fig:ddim-encode} we illustrate the differences between the $\bfx_T$ constructed from running the DiffAE forward solver, Equation (8) in~\cite{diffae}, versus simply sampling white noise.
Interestingly, a silhouette of the head is clearly visible in the encoded $\bfx_T$; moreover, there exists bands of relative uniformity emanating from this silhouette.
Clearly, the output of this forward solver is not the unit Gaussian we would expect for the formulation in~\cref{eq:sde-diffusion}.
However, as evidenced by the second row, this deviation from pure white noise is what enables the excellent reconstruction abilities of the Diffusion Autoencoder.
In the second row of~\cref{fig:ddim-encode} the output of the Diffusion Autoencoder model is shown.
As expected the image using the $\bfx_T$ from the DiffAE forward solver has a more faithful reconstruction whereas the image generated from white noise has noticeable variations.
The primary question is if these variations are enough to cause an FR system to reject the image.

In addition to investigating starting from pure white noise, we also examine only adding partial noise to the image.
From~\cref{eq:forward_eq} we can sample arbitrary levels of added noise by following the noise schedule, $\alpha_t, \sigma_t$.
Therefore, instead of starting the Diffusion model from $\bfx_T$, we could start from $\bfx_t$ for some $t < T$.
In~\cref{fig:dpm-partial-noise} we illustrate the morphing process. First we perform a pixel-wise average of the two aligned bona fide images.
We then inject noise in accordance to the noise schedule at time $t$ to get a noisy version of the pixel-wise morph.
The Diffusion model is then run from time $t$ back to $0$ to remove the added noise.
Note, the model is still conditioned on the morphed latent representation.
The goal is that the added noise can mask some of the artefacts from a pixel-wise average while retaining some of the low-frequency information that could be helpful to the generative process.

\begin{table}[h]
    \caption{Amount of added noise versus MMPMR ($\uparrow$) using the DPM++ 2M solver with $N = 50$.}
    \centering
    \begin{tabular}{lrrr}
    \toprule
     \textbf{Noise Level}                                & \textbf{AdaFace} &   \textbf{ArcFace} &  \textbf{ElasticFace} \\
    \midrule
    1.0                   &                4.5  &                3.48 &                    2.04 \\
    0.6                   &                9.41 &                6.75 &                    4.91 \\
    0.5                   &               15.13 &               12.27 &                    9.41 \\
    0.4                   &               27.61 &               21.68 &                   21.27 \\
    0.3                   &               45.81 &               40.49 &                   37.01 \\
    \bottomrule
    \end{tabular}
    \label{tab:mmpmr_noise}
\end{table}

In~\cref{tab:mmpmr_noise} we present the MMPMR values associated with this technique at various noise levels. We define the noise level to be $\frac tT$ for any given timestep $t$.
We discover that the forward ODE solver is paramount to the success of creating high quality Diffusion morphs, as evidenced by the abysmal MMPMR numbers resulting from white noise over the encoded $\bfx_T$.
We notice that as the noise level decreases the MMPMR does improve; however, we attribute this to the morphed images converging to the pixel-wise average rather than any merit to this particular idea.
While upon visual inspection we find images generated from white noise to have less artefacts in the morphed images and to retain more high-frequency details, we conclude that to create an effective morph using Diffusion Autoencoders, it is necessary to calculate $\bfx_T$ through some forward ODE solver.

\begin{table*}[t]
    \centering
    \caption{Study of the effects on autoencoding reconstruction quality across different forward ODE solvers on the FRLL dataset. }
    \begin{tabular}{lrrrrrrrr}
    \toprule
    &\multicolumn{4}{c}{\textbf{LPIPS($\downarrow$)}} & \multicolumn{4}{c}{\textbf{MSE($\downarrow$)}}\\
    \cmidrule(lr){2-5}\cmidrule(lr){6-9}
    \textbf{Forward ODE Solver} &
    $N_F = 20$ & $N_F = 50$ & $N_F = 100$ & $N_F = 250$&
    $N_F = 20$ & $N_F = 50$ & $N_F = 100$ & $N_F = 250$\\
    \midrule
    DiffAE,~\cref{eq:diffae_update_good} & \textbf{0.2370}&0.1404&0.1211&0.1113&\textbf{0.0037}&0.0020&0.0014&0.0010\\
    DDIM,~\cref{eq:forward_ddim}  & 0.2953&0.1843&0.1173&0.0760&0.0055&0.0015&\textbf{0.0005}&\textbf{0.0004}\\
    DPM++ 2M,~\cref{eq:forward_dpm_2m}  &0.3247&\textbf{0.1159}&\textbf{0.1120}&\textbf{0.0752}&0.0082&\textbf{0.0009}&\textbf{0.0005}&\textbf{0.0004}\\
    \bottomrule
    \end{tabular}
    \label{tab:ae_ode_comp}
\end{table*}

\subsection{Solving the Forward ODE}
\label{sec:forward_ode_solver}
Motivated by our findings in~\cref{sec:noise_inject} we decide to explore the forward ODE solver to see if we can achieve a reduction in NFE from this encoding.
The goal of the forward ODE solver is to find an $\bfx_T$ such that when used as the starting point for the Diffusion model the output is $\bfx_0$.
In order to discuss the forward ODE solver we briefly revisit the DDIM solver used to sample Diffusion models.
Using the conventions of Lu~\etal~\cite{lu2023dpmsolver} the original DDIM update equation can be written as follows
\begin{equation}
    \label{eq:ddim_update}
    \bfx_{t_{i-1}} = \frac{\sigma_{t_{i-1}}}{\sigma_{t_i}}\bfx_{t_i} - \alpha_{t_{i-1}}(e^{h_i} - 1)\bfx_\theta(\bfx_{t_i}, \bfz, t_{i})
\end{equation}
where $\bfx_\theta$ is the data prediction model which can be found from the noise prediction model via
\begin{equation}
    \bfx_\theta(\bfx_t, \bfz, t) = \frac{\bfx_t - \sigma_t\boldsymbol\epsilon_\theta(\bfx_t, \bfz, t)}{\alpha_t}
\end{equation}
and $h_i = \lambda_{t_i} - \lambda_{t_{i-1}}$ and $\lambda_t = \log \alpha_t - \log \sigma_t$ is the log Signal to Noise Ratio (log-SNR).
While~\cref{eq:ddim_update} provides a way to estimate $\bfx_0$ from $\bfx_T$, the goal of the forward ODE solver is to find $\bfx_T$ from $\bfx_0$, \ie, to move forward in time.
The forward update equation used by Diffusion Autoencoders and DiM can be found by rearranging~\cref{eq:ddim_update} to find the next sample $\bfx_{t_i}$ from $\bfx_{t_{i-1}}$.
\begin{equation}
    \label{eq:diffae_update_bad}
    \bfx_{t_i} = \frac{\sigma_{t_{i}}}{\sigma_{t_{i-1}}}\bigg(\bfx_{t_{i-1}} + \alpha_{t_{i-1}} (e^{h_{i}}-1)\bfx_\theta(\bfx_{t_{i}}, \bfz, t_{i})\bigg)
\end{equation}
However, this formulation clearly can't work for the forward pass as the calculation of $\bfx_{t_i}$ depends on an evaluation of the data prediction model on $\bfx_{t_i}$.
Preechakul~\etal~\cite{diffae} remedy this by evaluating the network on $\bfx_{t_{i-1}}$ instead, turning~\cref{eq:diffae_update_bad} into
\begin{equation}
    \label{eq:diffae_update_good}
    \bfx_{t_{i+1}} = \frac{\sigma_{t_{i+1}}}{\sigma_{t_{i}}}\bigg(\bfx_{t_{i}} + \alpha_{t_{i}} (e^{h_{i+1}}-1)\bfx_\theta(\bfx_{t_{i}}, \bfz, t_{i})\bigg)
\end{equation}
Note, for more consistent notation we wrote~\cref{eq:diffae_update_good} in terms of finding $\bfx_{t_{i+1}}$ from $\bfx_{t_i}$.

Doubtful of the validity substitution used to construct~\cref{eq:diffae_update_good} from~\cref{eq:diffae_update_bad}, we propose an alternative formulation to solving the forward ODE.
We observe that the aim of ``stochastic encoder'' from~\cite{blasingame2023leveraging,diffae} is quite similar to solving the PF-ODE,~\cref{eq:pf_ode}, as time runs forward from $0$ to $T$.
We propose to instead construct an additional ODE solver with the initial condition $\bfx_0$ that solves the PF-ODE forwards in time.
We propose two formulations: one using the first-order single-step DDIM solver, and the other using the second-order multi-step DPM++ 2M solver.

In Proposition 4.1 of~\cite{lu2023dpmsolver} Lu~\etal~show that an exact solution of the PF-ODE is given by
\begin{equation}
    \label{eq:exact_sol_pf_ode}
    \bfx_t = \frac{\sigma_t}{\sigma_s} \bfx_s + \sigma_t \int_{\lambda_s}^{\lambda_t} e^\lambda \bfx_\theta(\bfx_{\tau(\lambda)}, \tau(\lambda))\;\text d\lambda
\end{equation}
given some initial value $\bfx_s$ and where $\tau(\lambda) = t$ is a change of variables from time $t$ to log-SNR $\lambda$.
Setting $s = t_i$ and $t = t_{i+1}$ we construct a first-order approximation of~\cref{eq:exact_sol_pf_ode}
\begin{equation}
    \label{eq:forward_ddim}
    \bfx_{t_{i+1}} = \frac{\sigma_{t_{i+1}}}{\sigma_{t_i}}\bfx_{t_i} - \alpha_{t_{i+1}}(e^{-h_{i+1}} - 1)\bfx_\theta(\bfx_{t_i}, \bfz, t_{i})
\end{equation}
This first-order approximation of~\cref{eq:exact_sol_pf_ode} is the update equation of the DDIM solver as time runs forward, so we call it the DDIM solver for the forward ODE.

While at first glance this may appear similar to the DiffAE forward solver, we show that the local difference $\Delta$ at step $t_{i+1}$ between our solver and the DiffAE forward solver is given by the following equation
\begin{equation}
    \Delta = \bigg|\Big(\frac{\sigma_{t_{i+1}}}{\sigma_{t_i}}\alpha_{t_i}(e^{h_{i+1}}-1) - \alpha_{t_{i+1}}(e^{-h_{i+1}} - 1)\Big)\hat\bfx_\theta\bigg|
\end{equation}
where $\hat\bfx_\theta$ is the network evaluation at time $t_i$.
This reveals that while similar in goal, our strategy is meaningfully different from that in~\cite{diffae}.

\begin{figure*}[t]
    \centering
    \includegraphics[width=\textwidth]{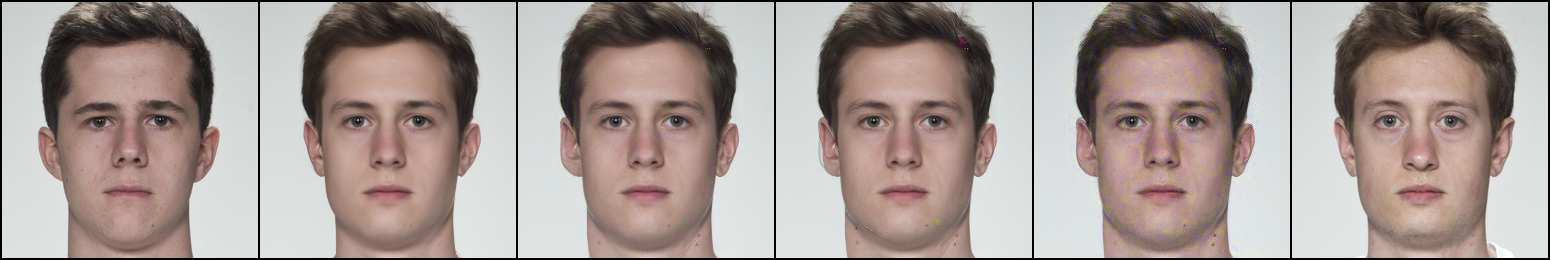}
    \caption{From left to right: identity $a$, DiffAE forward solver $N_F = 250$, DDIM forward ODE solver $N_F=100$, DPM++ 2M forward ODE solver $N_F=100$, DPM++ 2M forward ODE solver $N_F = 50$, and identity $b$.}
    \label{fig:forward_ode_comp}
\end{figure*}

Following the derivations of Lu~\etal~\cite{lu2023dpmsolver} we construct a second-order multi-step approximation of~\cref{eq:exact_sol_pf_ode} as time runs forward:
\begin{align}
    \label{eq:forward_dpm_2m}
    r_{i+1} &= \frac{h_i}{h_{i+1}}\nonumber\\
    \mathbf{D}_{i+1} &= \big(1 + \frac{1}{2r_{i+1}}\big)\bfx_\theta(\bfx_{t_i}, \bfz, t_i)\nonumber\\
                    &\quad- \frac{1}{2r_{i+1}}\bfx_\theta(\bfx_{t_{i-1}},\bfz, t_{i-1})\nonumber\\
    \bfx_{t_{i+1}}  &= \frac{\sigma_{t_{i+1}}}{\sigma_{t_i}}\bfx_{t_i} - \alpha_{t_{i+1}}\big(e^{-h_{i+1}} - 1\big)\mathbf{D}_{i+1}
\end{align}
This set of equations represent the heart of DPM++ 2M solver, hence we call this the DPM++ 2M solver for the forward ODE.

We believe that our formulations for solving the forward ODE are more principled than the one in~\cite{diffae} and we disagree with the validity of the substitution used to find~\cref{eq:diffae_update_good}.
To verify our theoretical intuition we experimentally compare our two formulations for the forward ODE solver against the DiffAE forward solver.
First, we assess the impact the forward ODE solver plays on the reconstruction ability of the Diffusion Autoencoder.
In~\cref{tab:ae_ode_comp} we measure the LPIPS metric and Mean Squared Error (MSE) between the original and reconstructed images from the FRLL dataset.
We use the same DPM++ 2M $N = 20$ PF-ODE solver across all three different forward ODE solvers.
We find that our proposed formulations vastly outperform the DiffAE forward solver in both LPIPS and MSE metrics.
Interestingly, the DiffAE forward solver performs best at $N_F = 20$ steps, however, the reconstruction error is too high across all three solvers to be useful for autoencoding or morphing applications.
We notice the DPM++ 2M forward ODE solver at $N_F = 50$ steps achieves almost the same performance as the DiffAE forward solver at $N_F = 250$ steps.
Our proposed implementation allows us to cut the NFE down by 200 to keep similar performance or down by 150 to achieve superior performance.

\begin{table}[h]
    \caption{Impact of forward ODE Solver on MMPMR.}
    \centering
    \begin{tabular}{lrrrr}
    \toprule
        && \multicolumn{3}{c}{\textbf{MMPMR}($\uparrow$)}\\
        \cmidrule(lr){3-5}
     \textbf{ODE Solver}                                 &   \textbf{NFE}($\downarrow$) & \textbf{AdaFace} &   \textbf{ArcFace} &  \textbf{ElasticFace} \\
        \midrule
        DiffAE  &250&               92.02 &               90.18 &                   93.05 \\
        DDIM&100 & 91.82 &               88.75 &                   91.21 \\
        DPM++ 2M &100&            90.59 &               87.12 &                   90.8  \\
        DDIM&50 & 89.78 &               86.3  &                   89.37\\
        DPM++ 2M &50&             90.18 &               86.5  &                   88.96 \\
    \bottomrule
    \end{tabular}
    \label{tab:mmpmr-forward-ode}
\end{table}

In addition, we evaluate the impact the choice of forward ODE solver has on morphing performance.
In~\cref{tab:mmpmr-forward-ode} we measure the MMPMR with different forward ODE solvers using the same DPM++ 2M PF-ODE solver with $N = 50$ steps for all experiments.
Note, we report the NFE only for solving the forward ODE.
Unfortunately, the superior autoencoding performance of the DDIM and DPM++ 2M ODE solvers does not seem to be reflected in the MMPMR numbers with the MMPMR experiencing a slight drop in performance across all three FR systems.
Interestingly, while DPM++ 2M slightly outperforms DDIM at $N_F = 50$ steps which aligns with our experimental observations in~\cref{tab:mmpmr-forward-ode}, we find that DDIM actually outperforms DPM++ 2M at $N_F = 100$.
In light of these results, we recommend the DDIM solver with $N_F = 100$ steps as the forward ODE solver, because it greatly reduces the NFE with only a slight decrease in MMPMR.

\cref{fig:forward_ode_comp} illustrates the impact on the morphing process the different forward ODE solvers play.
We notice that DDIM produces sharper and crisper images than DiffAE, preserving more of the high-frequency content of the original bona fide images. DPM++ 2M at $N_F = 100$ steps begins to create noticeable artefacts in the image which is only amplified when we reduced the number of steps to $N_F = 50$.
This visual assessment seems to roughly correlate with the performance observed in~\cref{tab:mmpmr-forward-ode}; however, our human assessment seems to favor the morphs using the DDIM $N_F = 100$ solver over the DiffAE solver.

\section{Results}
For completeness we compare our Fast-DiM algorithm against other face morphing algorithms on the SYN-MAD 2022 dataset. We compare against three landmark-based techniques: OpenCV, Webmorph, and FaceMorpher; in addition to the MIPGAN-I, MIPGAN-II, DiM-A, and DiM-C morphing attacks.
We denote our proposed model from~\cref{sec:ode_solver} as Fast-DiM and our proposed model from~\cref{sec:forward_ode_solver} as Fast-DiM-ode as it uses the forward ODE solver.
The Fast-DiM model uses $N = 50$ steps with the DPM++ 2M solver and the DiffAE forward solver with $N_F = 250$.
Likewise, the Fast-DiM-ode model uses the same PF-ODE solver, but the DDIM forward ODE solver with $N_F = 100$.

For DiM models we chose to report the total NFE across both solving the forward ODE and PF-ODE as $N + N_F$ rather than $N + 2N_F$, as one can simply batch the encoding of the two bona fide images into a single image tensor, exchanging time for memory.
We believe $N + N_F$ is a better representation of the NFE for these models as it represents the minimal NFE.
The MIPGAN family of models use 150 optimization steps when constructing the morphed image, so we report NFE = 150 for the MIPGAN models.

\begin{table}[h]
\caption{Vulnerability of different FR systems across different morphing attacks on the SYN-MAD 2022 dataset.}
\centering
\begin{tabular}{lrrrr}
\toprule
&& \multicolumn{3}{c}{\textbf{MMPMR}($\uparrow$)}\\
            \cmidrule(lr){3-5}
 \textbf{Morphing Attack}                                 & \textbf{NFE($\downarrow$)}  &\textbf{AdaFace} &   \textbf{ArcFace} &  \textbf{ElasticFace} \\
\midrule
 FaceMorpher~\cite{syn-mad22}                                     &-&               89.78 &               87.73 &                   89.57 \\
 Webmorph~\cite{syn-mad22}                                        &-&               97.96 &               96.93 &                   98.36 \\
 OpenCV~\cite{syn-mad22}                                          &-&               94.48 &               92.43 &                   94.27 \\
 MIPGAN-I~\cite{mipgan}                                      &150&               72.19 &               77.51 &                   66.46 \\
 MIPGAN-II~\cite{mipgan}                                       &150&               70.55 &               72.19 &                   65.24 \\
 DiM-A~\cite{blasingame2023leveraging}                                           &350&               92.23 &               90.18 &                   93.05 \\
 DiM-C~\cite{blasingame2023leveraging}                                           &350&               89.57 &               83.23 &                   86.3  \\
\textbf{Fast-DiM} &300&               92.02 &               90.18 &                   93.05 \\ 
 \textbf{Fast-DiM-ode}   &150&    91.82 &               88.75 &                   91.21    \\     
 \bottomrule
\end{tabular}
\label{tab:mmpmr}
\end{table}

\subsection{Vulnerability}
\cref{tab:mmpmr} provides the MMPMR at FMR = 0.1\% for all the evaluated morphing attacks and report the NFE, if applicable.
We notice that, unsurprisingly, the landmark-based morphing attacks are highly effective, often outshining their representation-based counterparts.
This trend has been noticed in prior works and is consistent with the current state of face morphing research~\cite{Blasingame2021LeveragingAL,mipgan, syn-mad22, blasingame2023leveraging}.
Fast-DiM represents only the slightest decline in performance from DiM-A on the AdaFace FR system but otherwise retains the excellent performance for a representation-based attack.
Fast-DiM-ode, however, experiences a decline in MMPMR of no more than 1.6\% while still maintaining superior performance to DiM-C and the MIPGAN models.
It even outperforms the landmark-based COTS FaceMorpher attack.
Fast-DiM, and Fast-DiM-ode, manage to outperform all other studied representation-based attacks with the exception of DiM-A.

\begin{table}[h]
    \centering
    \caption{MAP metric for all three FR systems on the SYN-MAD 2022 dataset.}
    \begin{tabular}{lrrrr}
    \toprule
    && \multicolumn{3}{c}{\textbf{Number of FR Systems}}\\
    \cmidrule(lr){3-5}
    \textbf{Morphing Attack} & \textbf{NFE($\downarrow$)} & \textbf{1} & \textbf{2} & \textbf{3}\\
    \midrule
     FaceMorpher~\cite{syn-mad22}                                     &- & 92.23 & 89.57 & 85.28 \\
     Webmorph~\cite{syn-mad22}                                        &-& 98.77 & 98.36 & 96.11 \\
     OpenCV~\cite{syn-mad22}                                          &-& 97.55 & 93.87 & 89.78 \\
     MIPGAN-I~\cite{mipgan}                                        &150& 85.07 & 72.39 & 58.69 \\
     MIPGAN-II~\cite{mipgan}                                       &150& 80.37 & 69.73 & 57.87 \\
     DiM-A~\cite{blasingame2023leveraging}                                           &350& 96.93 & 92.43 & 86.09 \\
     DiM-C~\cite{blasingame2023leveraging}                                           &350& 92.84 & 87.53 & 78.73 \\
     \textbf{Fast-DiM}   &300& 97.14 & 92.43 & 85.69 \\
     \textbf{Fast-DiM-ode}   &150& 95.91 & 91.21 & 84.66 \\
    \bottomrule
    \end{tabular}
    \label{tab:map}
\end{table}

We present the $\text{MAP}[1, c]$ values for the different morphing attacks in~\cref{tab:map}.
We observe that Fast-DiM achieves slightly higher performance in fooling a single FR system than DiM-A; however, it quickly loses out in the case of multiple FR systems.
The Fast-DiM and Fast-DiM-ode models outperform all representation-based morphing attacks other than DiM-A and even outperform the FaceMorpher attack.
Despite the dramatic reduction in NFE, the Fast-DiM and Fast-DiM-ode models manage to pose a potent threat to FR systems, managing to fool all three FR systems at least 84\% of the time.

\begin{table*}[t]
    \centering
    \caption{Detection Study on all training subsets of the SYN-MAD 2022 dataset.}
    \begin{tabular}{lrrrrrrrrrrrr}
    \toprule
     &\multicolumn{4}{c}{\textbf{Dataset-A}}
     &\multicolumn{4}{c}{\textbf{Dataset-B}}
     &\multicolumn{4}{c}{\textbf{Dataset-C}}\\
     \cmidrule(lr){2-5}
     \cmidrule(lr){6-9}
     \cmidrule(lr){10-13}
     &&\multicolumn{3}{c}{\textbf{APCER @ BPCER($\uparrow$)}}
     &&\multicolumn{3}{c}{\textbf{APCER @ BPCER($\uparrow$)}}
     &&\multicolumn{3}{c}{\textbf{APCER @ BPCER($\uparrow$)}}\\
     \cmidrule(lr){3-5}
     \cmidrule(lr){7-9}
     \cmidrule(lr){11-13}
     \textbf{Morphing Attack} &
     \textbf{EER($\uparrow$)} & \textbf{0.1\%} & \textbf{1.0\%} & \textbf{5.0\%} &
     \textbf{EER($\uparrow$)} & \textbf{0.1\%} & \textbf{1.0\%} & \textbf{5.0\%} &
     \textbf{EER($\uparrow$)} & \textbf{0.1\%} & \textbf{1.0\%} & \textbf{5.0\%}\\
    \midrule
 FaceMorpher~\cite{syn-mad22}                                     &  0    &                 0.49 &                 0    &                 0    & 91.67 &               100    &               100    &               100    &  5.39 &                46.08 &                16.18 &                 5.88 \\
 OpenCV~\cite{syn-mad22}                                         &  0    &                 0    &                 0    &                 0    & 46.57 &               100    &                97.55 &                89.71 &  0    &                 0    &                 0    &                 0    \\
 Webmorph~\cite{syn-mad22}                                        &  0    &                 0    &                 0    &                 0    & 46.57 &                99.51 &                98.04 &                91.18 &  0    &                 0    &                 0    &                 0    \\
 MIPGAN-I~\cite{mipgan}                                        &  2.94 &                17.65 &                 9.31 &                 1.47 &  0.98 &                 3.43 &                 0.98 &                 0.49 &  0    &                 0.49 &                 0    &                 0    \\
 MIPGAN-II~\cite{mipgan}                                       &  3.92 &                69.61 &                17.16 &                 2.45 &  0.98 &                 2.45 &                 0.98 &                 0.49 &  0    &                 0    &                 0    &                 0    \\
 DiM-A~\cite{blasingame2023leveraging}                                           & 34.31 &                96.57 &                91.18 &                77.94 & 75.49 &               100    &               100    &                99.02 &  0.98 &                 5.88 &                 1.47 &                 0    \\
 DiM-C~\cite{blasingame2023leveraging}                                           & 27.94 &                92.16 &                87.25 &                64.71 & 66.18 &               100    &                99.51 &                98.04 &  0    &                 0.49 &                 0    &                 0    \\
 \textbf{Fast-DiM} & 46.57 &                98.53 &                98.04 &                89.22 & 68.14 &               100    &               100    &                99.02 &  1.47 &                 5.88 &                 2.94 &                 0    \\
 \textbf{Fast-DiM-ode}   & 39.71 &                96.08 &                92.16 &                79.41 & 70.59 &               100    &               100    &                99.51 &  2.45 &                 7.35 &                 5.39 &                 0.49 \\
    \bottomrule
    \end{tabular}
    \label{tab:det_study}
\end{table*}

\subsection{Detectability Study}
To study the detectability of the Fast-DiM attacks, we implement an S-MAD detector trained on various morphing attacks.
We follow the approach of~\cite{blasingame2023leveraging} in designing our detectability study and use a SE-ResNeXt101-32x4d network pre-trained on the ImageNet dataset by NVIDIA as the backbone for our S-MAD detector.
The SE-ResNeXt101-32x4d is a state-of-the-art image recognition model based on the ResNeXt architecture with additional squeeze and excitation layers added.
We employ a stratified $k$-fold cross validation strategy in performing the detectability study to ensure fair reporting of the results and to preserve the class balance between morphed and bona fide images in each fold.
We opt to use $k = 5$ for our experiments.
We fine tune our detection model on three different subsets of the SYN-MAD 2022 dataset, each representing a different scenario for a potential S-MAD algorithm.
We enumerate these datasets as
\begin{enumerate}
    \item Dataset-A: consisting of FaceMorpher, OpenCV, and Webmorph.
    \item Dataset-B: consisting of MIPGAN-I and MIPGAN-II.
    \item Dataset-C: consisting of OpenCV, MIPGAN-II, and DiM-C morphs.
\end{enumerate}
We develop Dataset-A to illustrate an S-MAD algorithm trained on landmark-based attacks which may reflect an older S-MAD system.
We then develop Dataset-B to illustrate an S-MAD algorithm only trained on GAN-based attacks.
Lastly, we present Dataset-C to illustrate a realistic scenario for a strong S-MAD algorithm wherein the S-MAD algorithm is trained on a blend of different morphing attacks, one landmark-based, one GAN-based, and one Diffusion-based.
As we use a powerful pre-trained SE-ResNeXt101-32x4d model as our backbone, we only fine tune for 3 epochs and employ an exponential learning rate scheduler with differential learning rates to combat any potential overfitting of the model.
Additionally, we use a label smoothing with rate $0.15$ in our cross entropy loss function to further combat overfitting.
The S-MAD detection algorithm achieves a minimum of 98\% class balanced accuracy on each training fold before evaluating.
Importantly, with our $k$-fold strategy none of the bona fide images and morphs made from those bona fides used during training are used in evaluation.

To assess the detectability of the morphing attacks by the S-MAD system, we measure the Attack Presentation Classification Error Rate (APCER) and Bona fide Presentation Classification Error Rate (BPCER), in addition to the Equal Error Rate (EER).
In~\cref{tab:det_study} we present the results of the detectability study of the different morphing attacks evaluated against the S-MAD system trained on datasets, A, B, and C.
In Dataset-A we observe that just like the DiM models the Fast-DiM models are very difficult to detect when no DiM variant is present in the training algorithm, potentially posing a grave threat to pre-existing S-MAD systems.
Surprisingly, the MIPGAN variants are easily detected in this scenario as well even though the S-MAD model was only trained on landmark-based morphs.
In a similar vein, when training on the MIPGAN variants, Dataset-B, the performance of the S-MAD detector is abysmal with high error rates across the board with exception of the MIPGAN training set.
Interestingly, the FaceMorpher attack does exceedingly well in this scenario, which may be due in part to the uniqueness of the proprietary technique deployed by this morphing attack.
Lastly, in our most challenging scenario found in Dataset-C we observe that the inclusion of a DiM-C into the training set greatly reduced the effectiveness of all morphing attacks, showing that training on a diverse set of high quality morphing attacks is essential to achieving state-of-the-art S-MAD performance.
The FaceMorpher attack does better than rest in this scenario as well which we attribute to the same reasoning from before.
Both Fast-DiM variants have slightly higher error rates than their DiM counterparts.
We believe this is due to the difference in ODE solvers which we observed that give the images a sharper appearance compared to the more blurry appearance of the DiM models, see~\cref{fig:forward_ode_comp} for an illustration.

\section{Conclusion}
In this paper we have introduced Fast-DiM, an approach for generating high quality face morphs with lower NFE than existing models.
We have empirically demonstrated that our proposed model can use fewer NFE than previous Diffusion-based methods for face morphing while remaining a potent representation-based morphing attack.
We have shown that by replacing the DDIM PF-ODE solver with the DPM++ 2M PF-ODE solver in combination with solving the PF-ODE as time runs forwards using DDIM, we can achieve a remarkable reduction in NFE over prior methods.
Our results show that we can cut the NFE for solving the PF-ODE in half while retaining the same high quality morphing performance and that we can achieve an upwards of $85\%$ reduction in NFE for solving the PF-ODE as time runs forwards with only a maximal $1.6\%$ reduction in MMPMR.
We hope that this work will enable future exploration on techniques that leverage the iterative process of Diffusion models for face morphing that were once prohibitive due to the high computational demands of the previous methods.

\section*{Acknowledgment}
This material is based upon work supported by the Center for Identification Technology Research and National Science Foundation under Grant \#1650503.

\ifCLASSOPTIONcaptionsoff
  \newpage
\fi
\bibliographystyle{IEEEtran}
\bibliography{bib}

\end{document}